%% file: main.tex
\documentclass[letterpaper, 10 pt, conference]{ieeeconf}  %

\IEEEoverridecommandlockouts                              %

\usepackage{microtype}
\usepackage{graphicx}
\usepackage{url}
\usepackage[table]{xcolor}
\usepackage{subcaption}
\usepackage{booktabs}
\usepackage{float}
\usepackage{amsmath}
\usepackage{amssymb}
\usepackage{multirow}
\usepackage{makecell}
\usepackage{tabularx}
\usepackage{wrapfig}
\usepackage{adjustbox}
\usepackage{algorithm}
\usepackage{algorithmic}
\usepackage{graphicx}
\usepackage{xspace}
\usepackage{cuted}
\usepackage{tikz}
\usetikzlibrary{calc}

\makeatletter
\let\NAT@parse\undefined
\makeatother
\usepackage[numbers,sort&compress]{natbib}
\usepackage[hidelinks]{hyperref}

\newcommand\fakepar[1]{\noindent\textbf{#1}}

\title{\LARGE \bf
Simulation Priors for Rapid Adaptation in Model-Based RL\\for Autonomous Racing
}

\author{Adrian Remonda$^{1,2,\dagger}$, Jiajun Xi$^{2}$, Nicklas Hansen$^{2}$, Marcus Greiff$^{1}$, Thomas Lew$^{1}$,\\
John Talbot$^{1}$, John Subosits$^{1}$, and Xiaolong Wang$^{2}$%
\thanks{$^{1}$Toyota Research Institute, Los Altos, CA, USA.}%
\thanks{$^{2}$University of California San Diego (UCSD), La Jolla, CA, USA.}%
\thanks{$^{\dagger}$Work done in part during an internship at UCSD.}%
\thanks{Website: \url{toyotaresearchinstitute.github.io/pia-mbrl}}%
}

\begin{document}

\maketitle
\thispagestyle{empty}
\pagestyle{empty}

\newcommand\buffer{\mathcal{B}}
\newcommand\data{\mathcal{D}}
\newcommand\online[1]{#1_{\mathrm{on}}}
\newcommand\offline[1]{#1_{\mathrm{off}}}
\newcommand\datasetonline{$\online{\data}$\xspace}
\newcommand\datasetoffline{$\offline{\data}$\xspace}

\begin{strip}
    \centering
    \includegraphics[width=1\textwidth]
        {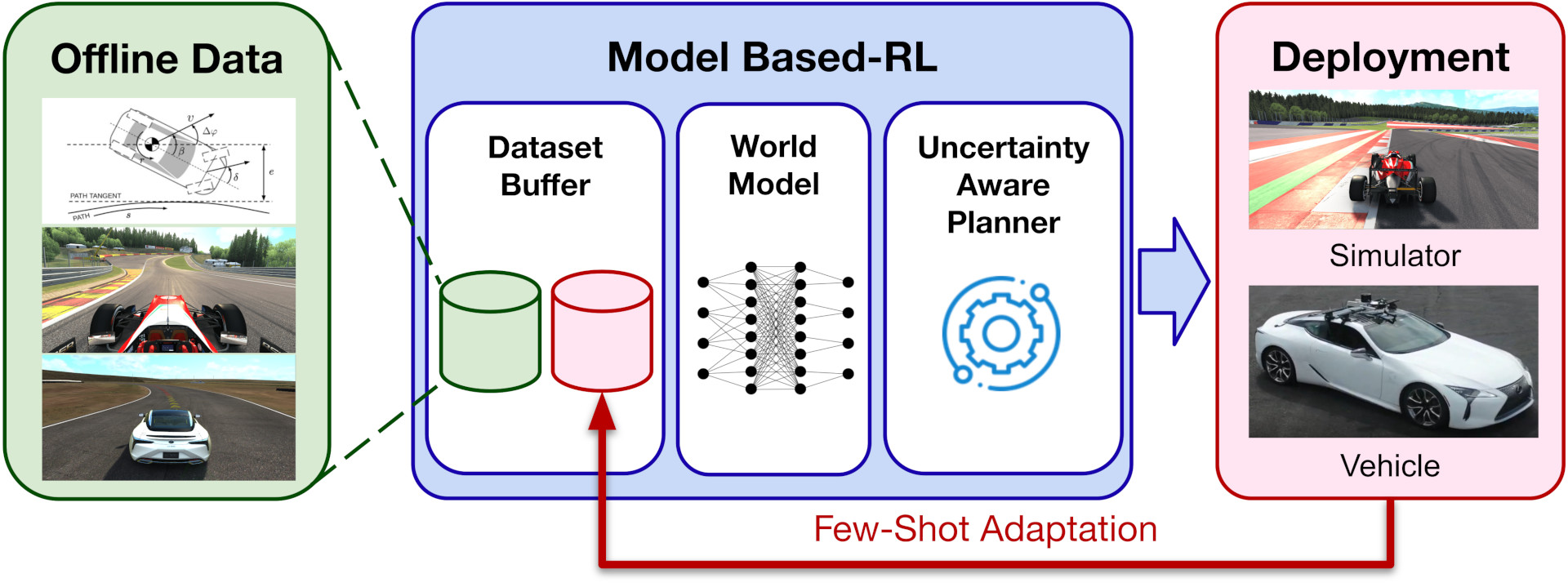}
    \captionof{figure}{%
        Prior-Informed Adaptation (PIA) pretrains a world model on heterogeneous simulation priors,
        then performs few-shot online adaptation in the target domain.
        Across racing tasks, PIA adapts faster than baselines
        and deploys on a vehicle despite sim-to-real mismatch.%
    }
    \label{fig:landing-diagram}
\end{strip}

\begin{abstract}

Reinforcement learning enables learning high-performance control policies, but its reliance on extensive interaction limits deployment on physical systems where data collection is costly, and failures are unsafe. We introduce Prior-Informed Adaptation (PIA), an offline-to-online model-based RL framework that uses structured simulation priors for rapid adaptation. PIA pretrains a world model on heterogeneous low- and high-fidelity simulated trajectories, then adapts it with limited target-domain interaction using uncertainty-regularized planning. Rather than requiring an accurate simulator for policy transfer, PIA treats simulated trajectories as biased priors whose residual mismatch is corrected during online adaptation. We evaluate PIA in autonomous racing using high-fidelity simulation and on a real full-size autonomous race car. In simulation, PIA adapts faster across track and surface shifts than strong baselines, achieving lower lap times and fewer failures. On hardware, PIA achieves reliable few-shot adaptation with less than one hour of on-vehicle interaction.
\end{abstract}

\section{INTRODUCTION}

Reinforcement learning (RL) has emerged as an effective data-driven framework for learning high-performance control policies, but its strongest successes often rely on large quantities of interaction data \cite{berner2019dota, schrittwieser2020mastering}. For example, \cite{Wurman2022} demonstrated that an RL policy can outrun the world's best human drivers in Gran Turismo, but only after $45{,}000$ hours of online interaction. This data requirement limits RL in settings where interaction is costly, failures are unsafe, or tasks change over time, such as robotics and real-world driving~\cite{lew2024riskaversemodelpredictivecontrol}.

As a result, RL practitioners often face a tradeoff between dataset diversity and interaction budget. Offline RL is attractive due to its reliance on existing data, but may lead to (unsafe) extrapolation errors when the policy encounters out-of-distribution states \cite{kumar2020conservativeqlearningofflinereinforcement, Kostrikov2021OfflineRL}. Such errors can be corrected with additional data collected by online RL \cite{nair2021awacacceleratingonlinereinforcement, Feng2023FinetuningOW, ball2023efficient, lee2021offlinetoonline}, but that too can be unsafe. In domains such as high-speed racing, where mistakes can be costly, this tradeoff often produces one of two undesirable outcomes: systems that require thousands of laps to achieve stable performance, which is unrealistic for real-world racing, or controllers that rely on highly accurate physics models, which limit flexibility and may potentially lead to suboptimal policies.
The challenge is to exploit structured but imperfect prior data while adapting to new geometries or dynamics under strict interaction budgets.

\begin{figure*}
    \centering
    \input{tikz/mosaic}
    \caption{Sequences from real-world deployments. After adaptation (bottom), PIA completes the oval racing task.}
    \label{fig:movingvehicle}
\end{figure*}
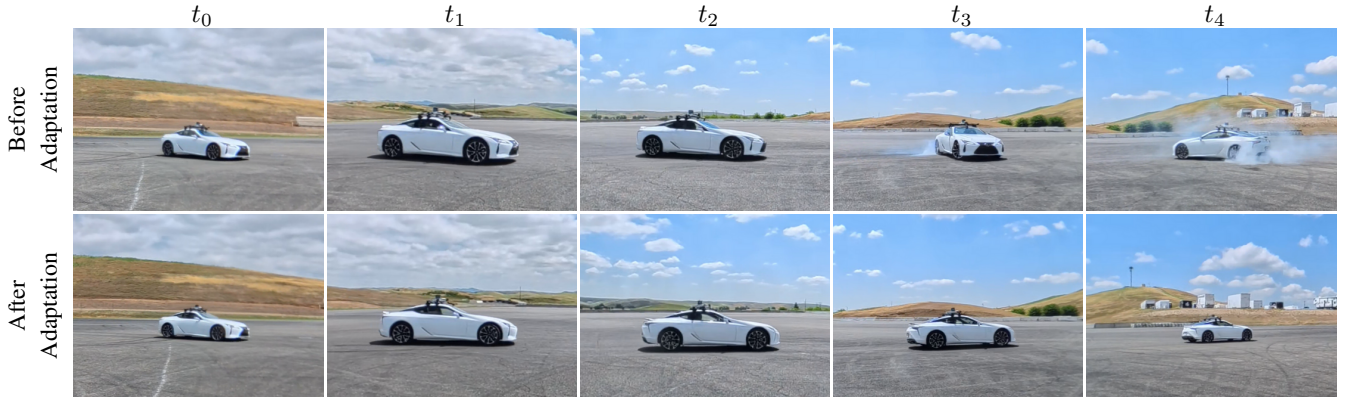

Prior work, such as Dyna-Q \cite{Sutton1990DynaAI, Janner2019WhenTT, Yu2020MOPOMO}, has shown that synthetic rollouts can significantly improve sample efficiency in model-based RL (MBRL) relative to model-free RL methods such as SAC \cite{Haarnoja2018SoftAA}. However, when rollouts are generated from learned dynamics, compounding prediction error can make long trajectories unreliable, which in practice restricts their usefulness to short rollouts. As a result, many MBRL methods rely on frequent replanning or conservative objectives to remain reliable. High-fidelity simulators used in control and robotics, such as MuJoCo \cite{Todorov2012MuJoCoAP} and NVIDIA Isaac Gym \cite{Makoviychuk2021IsaacGH}, provide a complementary source of synthetic experience: they can cheaply generate structured long-horizon trajectories, but depending on the simulator fidelity, may limit transfer to the target system. This motivates viewing data augmentation not simply as adding more data, but as \emph{simulation priors}: structured data generators that complement learned world models rather than replacing them.

In this work, we study how heterogeneous simulated trajectories can initialize modern MBRL when they do not exactly match deployment. We introduce Prior-Informed Adaptation (PIA), an offline-to-online MBRL framework that pretrains a world model on structured simulator data and then adapts using limited target-domain interaction. As shown in Figure~\ref{fig:landing-diagram}, a mixed replay retains broad prior coverage while updating the model with newly collected target data, and uncertainty-regularized planning discourages predictions associated with high epistemic uncertainty. Simulation is therefore used as an imperfect data prior rather than as an environment from which a policy transfers directly. This avoids the need for  an exact simulator or a domain-randomized distribution \cite{Peng_2018} that fully covers the target domain.

\begin{figure}
  \centering
  \includegraphics[width=\linewidth]{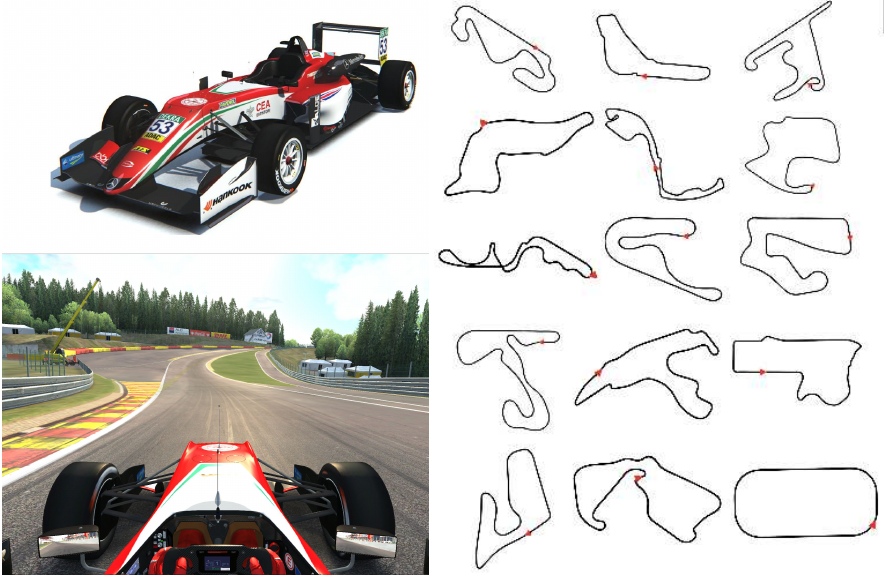}
  \vspace{-1.5em}
  \caption{\textbf{Tracks.} \emph{Left}: The F317 car and Assetto Corsa. \emph{Right}: Tracks considered in the ACgym experiments.}
  \label{fig:carsandtracks}
\end{figure}

We evaluate PIA in an autonomous racing task, where track geometry is typically available before deployment and can be simulated, while the vehicle dynamics remain difficult to reproduce accurately due to effects such as actuator latency, sensing/estimation mismatch, engine dynamics, changing weather, and tire variation. The Assetto Corsa racing simulator (ACGym) \cite{Remonda2024ASB}, shown in Figure~\ref{fig:carsandtracks}, provides a controlled high-fidelity proxy for real-world racing, allowing us to study rapid adaptation across unseen tracks and surface changes before hardware deployment. We then deploy the same offline-to-online pipeline on a full-size autonomous race car, where simulated trajectories are combined with a small amount of limited real hardware interaction to anchor the learned world model (see Figure~\ref{fig:movingvehicle}).
Extensive evaluations show that our proposed PIA method requires significantly fewer trials than other MBRL methods. 
Our main contributions are:
\begin{itemize}[
  \setlength{\labelindent}{0pt}
  \setlength{\labelsep}{0.4em}
]
  \item \textbf{A simulation-prior regime for offline-to-online MBRL}:
PIA uses heterogeneous %
trajectories as an imperfect offline initialization, then updates the learned world model with limited target-domain interaction rather than requiring direct policy transfer from an accurate simulator.
  \item \textbf{Adaptation rate and robustness in simulation}:
We conduct experiments in ACGym demonstrating that PIA improves adaptation rate, with lower lap times and fewer failures than offline-only and offline-to-online baselines.
  \item \textbf{Full-size autonomous racing deployment}:
To our knowledge, we provide the first demonstration of offline-to-online adaptation of a simulation-pretrained world model on a full-size autonomous race car, using less than one hour of real-car interaction.
  \item \textbf{Reproducible benchmark}:
We release code and datasets covering 15 tracks and multiple surface conditions, including more than {740 hours} of Assetto Corsa driving data and low-fidelity simulated rollouts, enabling future research on few-shot and sim-to-real adaptation.
\end{itemize}

Overall, our results show that structured simulated trajectories can substantially reduce the amount of target-domain interaction required for online adaptation.

\section{Related Work}
\label{sec:related_work}

\fakepar{Model-Based Augmentation.} Since Dyna-Q \cite{Sutton1990DynaAI}, synthetic rollouts from learned dynamics have been widely explored (e.g., MBPO \cite{Janner2019WhenTT}, MOPO \cite{Yu2020MOPOMO}, Dreamer \cite{Hafner2019DreamTC}, and TD-MPC2 \cite{Hansen2023TDMPC2SR}), and have shown that model-based rollouts can significantly improve sample efficiency. However, learned rollouts are prone to compounding error, so they are typically kept short to remain reliable. Physics-informed models such as PhIHP \cite{Asri2024PhysicsInformedMA} mitigate this by combining analytic priors with residual corrections, but are still constrained to short rollouts. AnyCar \cite{Xiao2024AnyCarTA} explores the use of transformers for world model generation and zero-shot deployment. While impressive, learned-model approaches remain tied to model fidelity when used for direct transfer.

\fakepar{Offline RL and Adaptation.}
Offline RL methods such as CQL and IQL 
\cite{levine2020offlinereinforcementlearningtutorial,kumar2020conservativeqlearningofflinereinforcement,Kostrikov2021OfflineRL}
mitigate extrapolation error through conservative or implicit constraints on value learning, but this can limit generalization. H2O \cite{Niu2022WhenTT}  combines a fixed offline real-world dataset with online rollouts from a simulator using dynamics-aware value regularization and corrected Bellman updates. PIA considers the complementary direction: it pretrains from fixed simulator trajectories and adapts through limited online interaction with the target environment.
\cite{nair2021awacacceleratingonlinereinforcement,lee2021offlinetoonline,ball2023efficient}.
FOWM \cite{Feng2023FinetuningOW} pretrains a world model offline on target-domain data and finetunes it with limited online interaction using uncertainty-regularized planning.
To mitigate distribution shift when learning, FOWM augments TD-MPC with a V-AWR objective that biases learning toward actions supported by the offline data. PIA follows the same offline-to-online template, but targets a different transfer setting: its offline initialization is constructed from heterogeneous simulation priors. Rather than using the V-AWR offline objective, PIA retains the TD-MPC2 training objective and conditions the model on task context to distinguish heterogeneous dynamics. Thus, the central problem is not only offline-to-online fine-tuning under task or policy shift, but cross-domain correction of the simulator's bias with limited target-domain interaction.

\fakepar{Sim-to-real Adaptation.} Domain randomization trains policies over randomized simulator dynamics to improve transfer under simulator mismatch \cite{Peng_2018}.
Rapid motor adaptation instead trains a base policy and adaptation module in simulation, enabling fast online adaptation to unseen deployment conditions \cite{kumar2021rmarapidmotoradaptation}.
These methods are most effective when the simulated variation covers the relevant target-domain variation.
More broadly, sim-to-real systems may incorporate real data for system identification, residual modeling, or validation; Kaufmann et al. \cite{Kaufmann2023ChampionLevelDroneRacing} improve drone racing by augmenting simulation with data-driven residual perception and dynamics models identified from real-world experience.
PIA is complementary: simulated trajectories are treated as biased replay priors rather than as the final training environment, and their residual mismatch is corrected with target-domain interaction during offline-to-online MBRL.

\fakepar{Autonomous Racing.} Model Predictive Control (MPC) delivers strong performance on full-size autonomous race cars but requires carefully engineered models \cite{tum, raji1, raji3, wischnewski2, lew2024riskaversemodelpredictivecontrol}. At the limits of handling, however, small residual mismatches in tire grip or actuator response can cause failures, motivating controllers that can update from physical interaction~\cite{davydov2025first}. RL systems have achieved strong autonomous-racing results in simulation with large interaction budgets, including GT Sophy in Gran Turismo \cite{Wurman2022, Fuchs_2021}, Formula RL \cite{remonda2022formularldeepreinforcement}, and AssettoCorsaGym \cite{Remonda2024ASB}, but these works do not address few-shot physical adaptation on a full-size race car. Closest to our hardware setting, Djeumou et al. \cite{Djeumou2024ReferenceFreeFD} train RL policies in a neural-SDE vehicle model learned from real driving data and deploy them zero-shot for a drifting task. PIA instead adapts a simulation-pretrained world model using limited physical interaction. 
To our knowledge, none of the cited literature has demonstrated this offline-to-online adaptation regime on a full-size autonomous race car.

\section{Preliminaries}
\label{sec:preliminaries}

In this section, we introduce notation and terminology that will appear throughout the paper. We formulate the autonomous racing problem as a Partially Observable Markov Decision Process (POMDP) $\langle \mathcal{S}, \mathcal{A}, \mathcal{O}, \mathcal{T}, r, \gamma \rangle$, where \(\mathcal{S}\) denotes the state space, \(\mathcal{A}\) is the action space, $\mathcal{O}$ is the observation space, $\mathcal{T} \colon \mathcal{S} \times \mathcal{A} \mapsto \mathcal{S}$ is the unknown transition model, $r \colon \mathcal{S} \times \mathcal{A} \mapsto \mathbb{R}$ is the reward function, and \(\gamma \in [0, 1)\) is the discount factor~\cite{Bellman1957MDP, kaelbling1998pomdp}. We approximate states as a sequence of the $k$ most recent observations. A policy $\pi$ selects actions ${a}_{t} \sim \pi(\cdot | {s}_{t})$ at each time step $t$ to maximize the expected discounted return \(\mathbb{E}_\pi \left[ \sum_{t=0}^\infty \gamma^t r({s}_{t}, {a}_{t}) \right]\). We assume that trajectories from the source and target environment share a compatible action space and observation interface, or can be mapped into one.

\fakepar{Model-Based RL.} In MBRL, the agent learns a predictive model of environment dynamics and leverages it for planning. This improves sample efficiency over model-free methods and enables sim-to-real transfer. One such example is TD-MPC2~\cite{Hansen2023TDMPC2SR}, which learns a latent world model from sequential data and performs short-horizon planning with MPC. Its end-to-end training and latent-space planning make it well-suited for multitask learning and diverse datasets.
Its world model consists of an encoder $h_\theta$, latent dynamics $d_\theta$, reward predictor $R_\theta$, and value predictor $Q_\theta$. 
Given trajectory segments sampled from the replay buffer $\mathcal{B}$, these components are jointly optimized according to
\begin{equation}
\mathcal{L}(\theta)
=
\mathbb{E}_{\tau \sim p_{\mathcal{B}}}
\left[
\ell_{\mathrm{TD\text{-}MPC2}}(\theta;\tau)
\right],
\label{eq:tdmpc2_loss}
\end{equation}
where
$\tau=\{(s_t,a_t,r_t,s_{t+1})\}_{t=0}^{T-1}$
is a trajectory segment of length $T$ sampled from the replay distribution
$p_{\mathcal{B}}$, and
$\ell_{\mathrm{TD\text{-}MPC2}}$ comprises latent consistency, reward prediction, and temporal-difference value prediction losses.
The policy prior $\pi_\theta$ is optimized separately from $(h_\theta, d_\theta, R_\theta,Q_\theta)$. For a candidate action sequence $a_{0:H-1}$, latent dynamics
$z_{h+1}=d_\theta(z_h,a_h)$ are predicted over a horizon
$h=0,\ldots,H-1$, with $z_0=h_\theta(s_t)$.
The planner scores each rollout using predicted rewards and a terminal value
$Q_\theta(z_H,a_H)$, where $a_H=\pi_\theta(z_H)$ is provided by the learned policy prior.

\fakepar{Simulation Priors.} We study rapid adaptation in a target environment where online interaction is limited, and failures may be costly. Before adaptation, the agent has access to offline trajectories from source simulators that provide useful structure and coverage, but are not assumed to match the target dynamics. We refer to these sources as \emph{simulation priors}. The goal is to use these biased but structured trajectories to enable more rapid adaptation with limited target-domain interaction.

\fakepar{Offline RL and Adaptation.} Given a dataset \datasetoffline, offline RL seeks to learn $\pi$ without additional interaction~\cite{levine2020offlinereinforcementlearningtutorial}. The main challenge is out-of-distribution generalization when $\pi$ reaches unseen states. A common strategy to mitigate this issue is offline pretraining followed by online adaptation, where a world model is initialized on \datasetoffline and refined with limited new interaction dataset \datasetonline (see~\cite{Feng2023FinetuningOW}).

\fakepar{ODE Model.} For low-fidelity simulations, we use a simple vehicle model described by ordinary differential equations $\dot{\mathbf{x}} = g(\mathbf{x}, \mathbf{u})$, where $\mathbf{x}$ is the vehicle state, $\mathbf{u}$ the control inputs. We use a dynamic bicycle model with aerodynamics loads, drivetrain dynamics, and tire forces. \cite{raji2}

\fakepar{Assetto Corsa Gym.} For high-fidelity simulation, we use ACGym \cite{Remonda2024ASB}, an open-source framework built on the Assetto Corsa simulator. It provides more realistic vehicle dynamics and geometries across several cars and laser-scanned tracks.

\vspace{-0.05in}
\section{PIA: Prior-Informed Adaptation} %
\label{sec:method}
\vspace{-0.05in}
Simulation data is cheap to generate, whereas real-world data is costly and potentially dangerous to collect. This motivates using simulation as a prior. PIA treats simulation priors as useful but mismatched data for initializing an MBRL world model: simulated trajectories supply broad structured coverage, limited target-domain interaction corrects their residual bias, and uncertainty-regularized planning keeps control conservative while that correction is still incomplete. %
PIA has two components: %
\begin{enumerate}
\item\emph{mixed replay, context-aware, offline-to-online training}, which updates the world model from a mixture of simulation-prior and target-domain transitions;
\item \emph{uncertainty-regularized latent planning}, which discourages the controller from exploiting predictions in regions poorly supported by target-domain data 
\end{enumerate}
While each component has prior work, their combination enables PIA's efficient offline-to-online performance.

\fakepar{Mixed Replay, Context-Aware, Offline-to-Online Training.}
\label{sec:mixed_replay}
First, PIA trains the policy from replay buffers containing heterogeneous trajectories from multiple sources.
In our experiments, the simulation-prior data consist of low-fidelity ODE rollouts and high-fidelity Assetto Corsa trajectories.

Second, all data is mapped to a common transition format and labeled with task and source context, denoted by $c$, to distinguish data sources such as ODE rollouts, Assetto Corsa simulation, and real-car interaction, as well as the corresponding track or deployment setting. When a channel is unavailable in one source, we estimate it from available variables when possible, and otherwise set it to zero. 

Third, PIA consists of training on a mixed replay buffer containing offline and online data (see Alg.~\ref{alg:pia}).  The offline buffer $\offline{\buffer}$ contains simulation-prior trajectories and any target-domain data available before adaptation. The online buffer $\online{\buffer}$ contains target-domain data collected online and used for adaptation.
After collecting an episode, PIA samples from a mixture %
$q_\alpha=\alpha \offline{p}
+
(1-\alpha)\online{p}$, where  
 $\offline{p}$ and $\online{p}$ denote uniform distributions over $\offline{\buffer}$ and $\online{\buffer}$ \cite{lee2021offlinetoonline, ball2023efficient, Feng2023FinetuningOW}. 
\begin{algorithm}[t]
  \footnotesize
  \caption{PIA: Offline-to-Online Adaptation}
  \label{alg:pia}
  \begin{algorithmic}
    \STATE \textbf{Inputs:} offline data $\offline{\data}$, target environment $E$, pretraining steps $N_{\mathrm{pre}}$, online episodes $N_{\mathrm{ep}}$, batch size $b$, replay ratio $\alpha$
    \STATE $\offline{\buffer} \leftarrow \offline{\data}$; \quad $\online{\buffer} \leftarrow \emptyset$
    \STATE {\color{gray}\textit{Offline pretraining}}
    \FOR{$t = 1$ \textbf{to} $N_{\mathrm{pre}}$}
      \STATE $\buffer \leftarrow \mathrm{sample}(\offline{\buffer}, b)$
      \STATE Update $\theta$ by minimizing $\mathcal{L}_\alpha(\theta)$ over $\mathcal{B}$
    \ENDFOR

    \STATE {\color{gray}\textit{Online Adaptation}}
    \FOR{$e = 1$ \textbf{to} $N_{\mathrm{ep}}$}
      \STATE Sample $\tau_e=\{(s_t,a_t,r_t,s_{t+1},c)\}_{t=0}^{T_e-1}$ from $E$
      \STATE $\alpha\leftarrow \alpha-(2N_{\mathrm{ep}})^{-1}$
      \STATE $\online{\buffer}  \leftarrow \online{\buffer} \cup \{\tau_e\}$
      \STATE $\buffer \leftarrow \mathrm{sample}(\offline{\buffer}, \lfloor \alpha b \rfloor) \cup \mathrm{sample}(\online{\buffer} , b-\lfloor \alpha b \rfloor)$
      \STATE Update $\theta$ by minimizing $\mathcal{L}_\alpha(\theta)$ over $\buffer$
    \ENDFOR
  \end{algorithmic}
\end{algorithm}
The TD-MPC2 model is trained on context-aware trajectory segments sampled
from the mixed replay distribution,
\begin{equation}
\mathcal{L}_{\alpha}(\theta)
=
\mathbb{E}_{(\tau,c)\sim q_{\alpha}}
\left[
\ell_{\mathrm{TD\text{-}MPC2}}(\theta;\tau,c)
\right],
\label{eq:mixed_replay_loss}
\end{equation}
where $c$ denotes the context associated with trajectory segment $\tau$, encoding its task and data source.
The policy prior is optimized separately as in TD-MPC2, but with additional context conditioning.
For pretraining, we let $\alpha=1$, which recovers the TD-MPC2 formulation. For online adaptation, we anneal $\alpha$ from $1$ to $0.5$ following a linear decay schedule. 
This combination of simulation-prior data and target-domain data ensures training a model that benefits from both high data coverage and limited but high-quality target-domain data.

\fakepar{Uncertainty-Regularized Latent Planning.}
\label{sec:uncertainty_planning}
While collecting target-domain data, the agent must act while this corrective data is still incomplete. 
Early on during adaptation, value function estimates may reflect simulator-prior bias.
To mitigate this, PIA uses FOWM-style uncertainty-regularized latent planning, with epistemic uncertainty estimated from Q-ensemble disagreement \cite{Feng2023FinetuningOW}. We extend the planner to be \emph{context-aware}. Given state $s_t$ and context $c$ the encoder becomes $z_0=h_\theta(s_t,c)$ and the terminal action is $a_H=\pi_\theta(z_H,c)$. 
Following \cite{Feng2023FinetuningOW}, define $u_h = \operatorname{Std}(\{Q_\theta^{(i)}(z_h,a_h,c)\}_{i=1}^{K})$,
where $\bar Q_\theta(z_h,a_h,c)=\frac{1}{K}\sum_{i=1}^{K}Q_\theta^{(i)}(z_h,a_h,c)$ denotes the ensemble mean. The planner scores each rollout by predicted rewards, terminal value, and uncertainty,
\begin{equation}
\begin{aligned}
G_\theta(z_0,a_{0:H-1},c)
={}& \sum_{h=0}^{H-1}\gamma^h
\left(
\hat r_\theta(z_h,a_h,c)-\lambda u_h
\right) 
\\
&+\gamma^H
\left(
\bar Q_\theta(z_H,a_H,c)-\lambda u_H
\right).
\end{aligned}
\label{eq:regularized_planning_score}
\end{equation}
where $\lambda$ is the uncertainty penalty coefficient.

\section{Experiments in Simulation}

First, we extensively evaluate PIA for autonomous racing in simulation using Assetto Corsa as the high-fidelity simulator and the simple ODE vehicle model as the low-fidelity simulation prior. We test whether simulation-prior pretraining improves performance and safety on unseen tracks and remains robust under surface-grip shifts.

\begin{table*}[t]
\centering
\begin{minipage}[t]{0.52\linewidth}
\centering
\caption{\textbf{Performance.} Best lap time (s). IQL, SAC scratch, FOWM, PIA: 3 seeds. Avg. only if all tasks completed.}
\label{tab:e1_aggregated_results}
\small
\setlength{\tabcolsep}{1.8pt}
\begin{tabular}{lccccc}
\toprule
Task   & IQL & SAC & FOWM & PIA-AC & PIA-ODE \\
\midrule
IND              & 159.33 & 60.00  & 60.19  & \textbf{59.97} & 60.00 \\
RBR              & Fail   & 255.64 & 87.83  & 85.85  & \textbf{84.34} \\
SIL              & 209.95 & Fail   & 118.88 & 114.16 & \textbf{112.17} \\
SIL$_\text{varying}$ & Fail   & Fail   & 122.82 & 114.94 & \textbf{112.81} \\
SIL$_\text{low}$ & Fail   & Fail   & 125.85 & 119.79 & \textbf{117.61} \\
\midrule
Average          & --     & --     & 103.11 & 98.94  & \textbf{97.39} \\
\bottomrule
\end{tabular}
\end{minipage}\hfill
\begin{minipage}[t]{0.45\linewidth}
\centering
\caption{\textbf{Miles per failures.} Results across tracks for SAC scratch, FOWM, and PIA. 3 seeds. {Higher is better.}}
\label{tab:e1_aggregated_results_safety}
\small
\setlength{\tabcolsep}{2.5pt}
\begin{tabular}{lcccc}
\toprule
Task    & SAC & FOWM & PIA-AC & PIA-ODE \\
\midrule
IND              & 80.73 & 714.68 & 1194.29 & \textbf{1233.25} \\
RBR              & 0.39  & \textbf{39.14} & 26.40 & 34.48 \\
SIL              & 0.28  & 32.06 & 34.62 & \textbf{68.80} \\
SIL$_\text{varying}$ & 0.02 & 19.54 & 42.68 & \textbf{46.57} \\
SIL$_\text{low}$ & 0.02 & 16.50 & 16.48 & \textbf{25.57} \\
\midrule
Average          & 16.29 & 164.38 & 262.09 & \textbf{281.73} \\
\bottomrule
\end{tabular}
\end{minipage}
\vspace{-.15in}

\end{table*}

\begin{figure*}
    \centering
    \includegraphics[width=\linewidth]{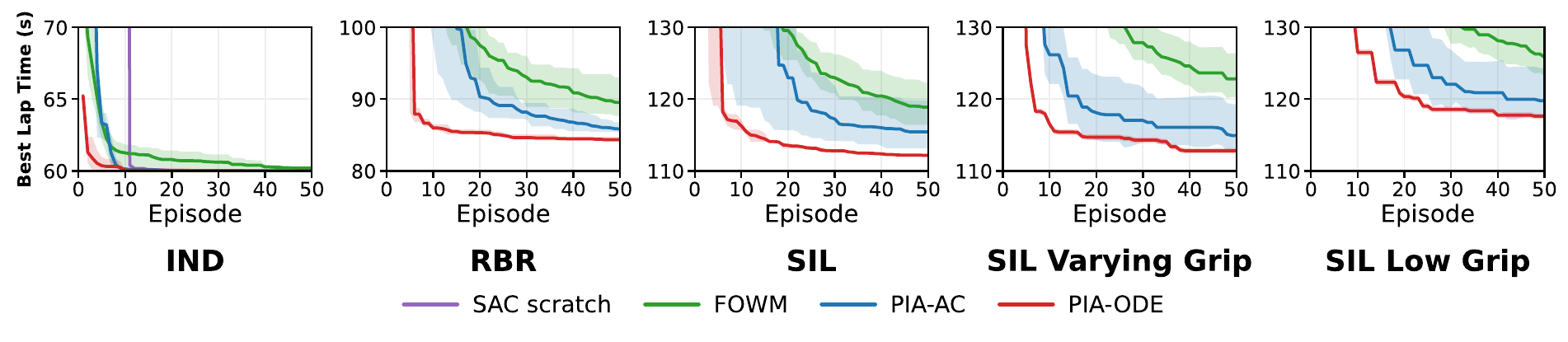}
    \vspace{-14pt}
    
    \caption{\textbf{Adaptation performance.} Best valid lap time by episode. Mean and $\pm 1$ standard deviation over three seeds.}
    \label{fig:e1_laptimes}
    \vspace{-0.15in}
\end{figure*}

We evaluate in ACGym on 9 tracks total: 6 Assetto Corsa training tracks and three held-out test tracks, Indianapolis (IND), Red Bull Ring (RBR), and Silverstone (SIL). We distinguish two settings. PIA-AC is pretrained only on AC trajectories from source tracks, and therefore tests transfer to held-out track geometry. PIA-ODE additionally receives low-fidelity ODE trajectories generated on the deployment-track geometry, but no high-fidelity AC trajectories from that track; it therefore tests cross-fidelity adaptation with known geometry and mismatched target dynamics. Surface experiments keep the SIL geometry fixed and shift grip among \textit{optimal}, \textit{varying}, and \textit{low} grip conditions. Episodes last up to 15k environment steps and terminate early if the car stalls or leaves the track. We report the best lap time. Safety and stability are captured by the number of failures and miles per failure. An incident is an off-track excursion or low-speed stall, and miles per incident is the total driving distance divided by the number of incidents.

\fakepar{Datasets.} We construct two sources of pretraining data. \emph{(1) AC (high-fidelity).} We train SAC from scratch for 6M steps per track in ACGym to generate expert-level trajectories. These datasets cover all training splits (1T--6T). \emph{(2) ODE (low-fidelity).} We introduce a lightweight ODE environment that shares the same observation and action interface as ACGym. We run the same SAC agent used for AC, but replace the physics step with the ODE model while preserving geometry and localization signals. The target-track ODE trajectories are intentional prior information in PIA-ODE; they provide geometry and approximate dynamics. PIA-ODE is a known-geometry cross-fidelity setting, whereas PIA-AC is the held-out-geometry setting.

\fakepar{Baselines.} We compare online SAC \cite{Haarnoja2018SoftAA}, offline IQL \cite{Kostrikov2021OfflineRL}, offline-to-online FOWM \cite{Feng2023FinetuningOW}, and our PIA variants, PIA-AC and PIA-ODE. FOWM and PIA-AC use identical offline buffers and update budgets. We include FOWM as an end-to-end baseline using its original TD-MPC formulation. Because FOWM and PIA differ in both learning objective and backbone, this comparison measures system-level performance rather than isolating an individual component. We evaluate two variants. PIA-AC is pretrained on six AC training tracks, and tests transfer to tracks absent from pretraining. PIA-ODE uses the same AC data and adds ODE rollouts on the three tracks held out from AC pretraining, matching realistic racing workflows in which the circuit is rehearsed in simulation before deployment while the target dynamics are uncertain.

\paragraph{Adaptation}
Tables~\ref{tab:e1_aggregated_results} and~\ref{tab:e1_aggregated_results_safety} show that IQL and SAC struggle on unseen tasks, while FOWM adapts but remains slower and less safe than PIA. PIA-ODE achieves the best average lap time and miles per incident. The improvements are most evident on complex tracks and low-grip surfaces. Even on simpler tasks such as IND with small lap-time gains, the low-fidelity ODE data augmentation improves safety by increasing the miles per failure metric.

\paragraph{Robustness to Domain Shift}
To assess robustness, we fine-tune on \textit{varying} and \textit{low} grip surfaces on the SIL track. These conditions change grip, while ODE rollouts use nominal vehicle parameters without surface-specific friction variation. This is critical in racing, where grip varies with weather, and controllers that overfit to a single surface fail when conditions change. As shown in Tables~\ref{tab:e1_aggregated_results_safety} and~\ref{tab:e1_aggregated_results} and Figure~\ref{fig:e1_laptimes}, all methods degrade relative to optimal-surface evaluation; SAC fails, while PIA-ODE maintains lower lap times and higher miles per incident than AC-only and FOWM on both low-grip variants. This suggests that nominal ODE rollouts provide transferable structure, although varying ODE friction parameters may further improve robustness.

\paragraph{Additional analyses and ablations}
Appendix~\ref{sec:appendix:ode_signals} separates the roles of ODE and AC data. The ODE-signal ablation in Appendix~\ref{sec:appendix:synthetic_only} directly addresses the known-geometry setting: retaining only explicit geometry channels improves safety but not adaptation speed, while masking those channels and retaining dynamics signals does not improve performance; combining both produces the strongest result. Removing uncertainty regularization leads to substantially worse safety, and task and context identifiers improve generalization when mixing heterogeneous data. %

\section{Vehicle Deployment}
\label{sec:real_car}

\newcommand\captionline[1]{\protect\tikz[baseline=-0.0ex]{\protect\draw[black, ultra thick] (-0.02,0.1) -- (0.32,0.1);\protect\draw[#1, thick] (0,0.1) -- (0.3,0.1);}}

\begin{wrapfigure}{r}{0.28\textwidth}
    \centering
    \includegraphics[width=\linewidth]{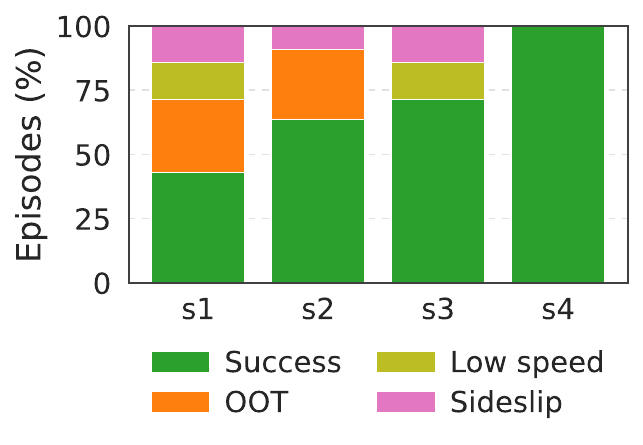}
    \caption{\textbf{Episode outcomes.} Distribution of successful episodes and failures during real-car adaptation.}
    \label{fig:real_episode_outcome_counts}
\end{wrapfigure}

\begin{figure}[t]
    \centering
    \begin{subfigure}[t]{0.245\linewidth}
        \centering
        \includegraphics[width=\linewidth]{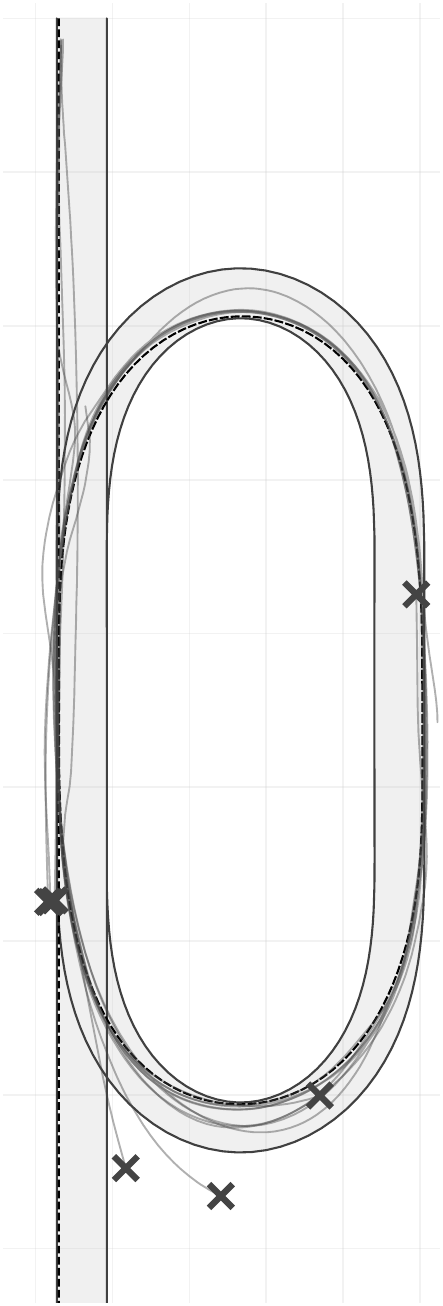}
        \caption{\texttt{s1}}
    \end{subfigure}\hfill
    \begin{subfigure}[t]{0.245\linewidth}
        \centering
        \includegraphics[width=\linewidth]{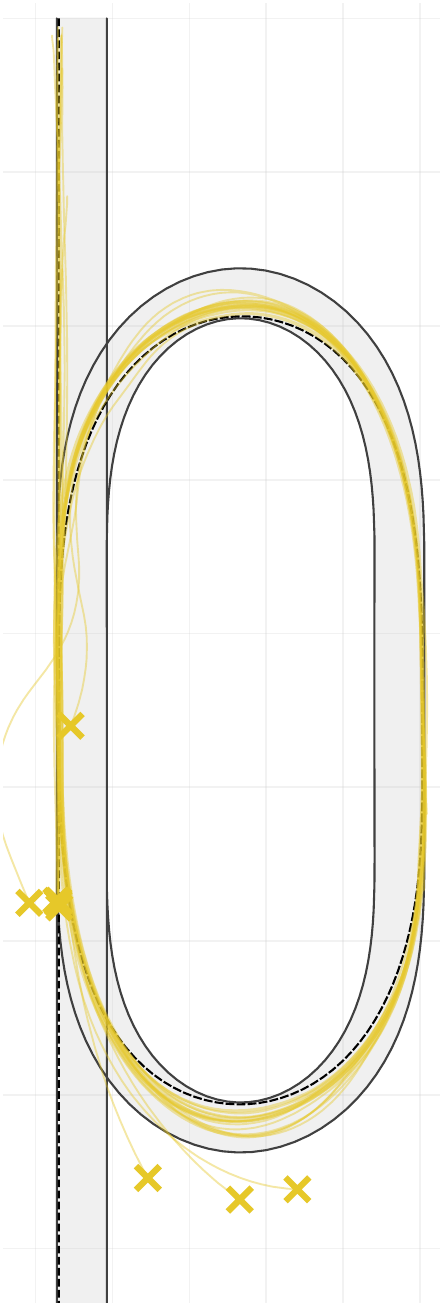}
        \caption{\texttt{s2}}
    \end{subfigure}\hfill
    \begin{subfigure}[t]{0.245\linewidth}
        \centering
        \includegraphics[width=\linewidth]{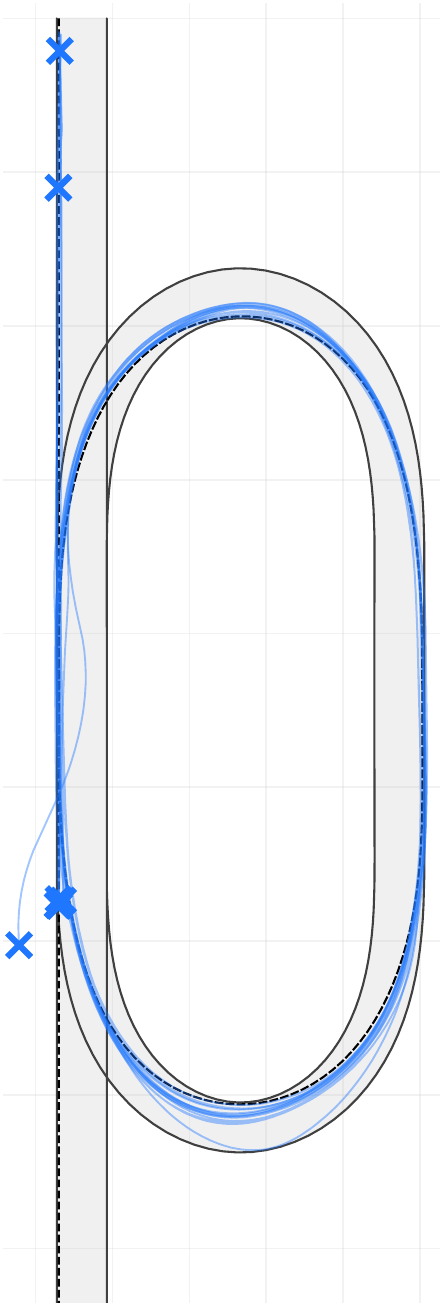}
        \caption{\texttt{s3}}
    \end{subfigure}\hfill
    \begin{subfigure}[t]{0.245\linewidth}
        \centering
        \includegraphics[width=\linewidth]{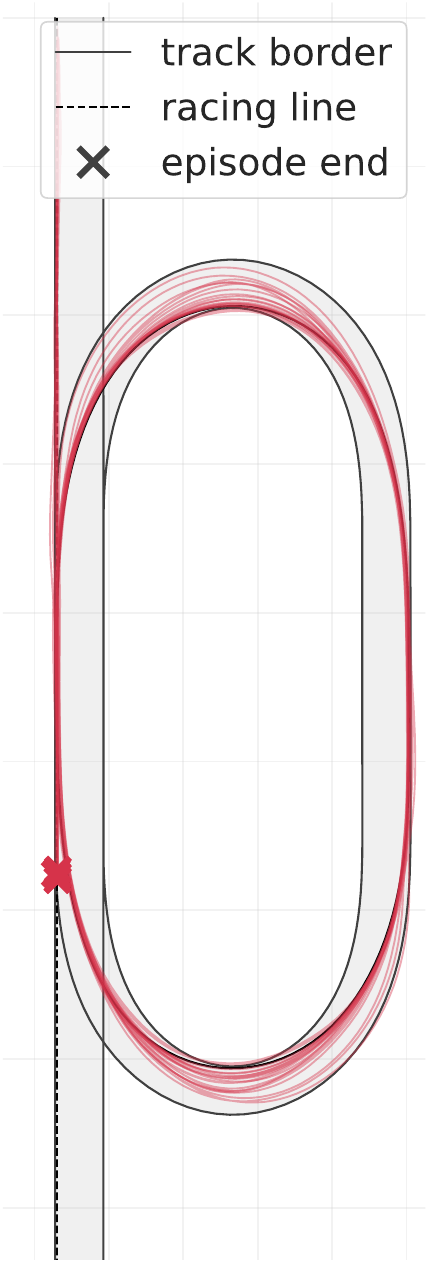}
        \caption{\texttt{s4}}
    \end{subfigure}
    \caption{\textbf{Vehicle XY traces.} All episode trajectories for deployment sessions \texttt{s1}--\texttt{s4}.}
    \vspace{-5mm}
    \label{fig:real_xy_traces_all_episodes}
\end{figure}

The previous experiments isolate the role of augmentation in a controlled sim-to-sim transfer setting.  We now evaluate the same adaptation pipeline on a full-size autonomous race car. We conduct a progressive real-car adaptation study on an oval race track (see Figure~\ref{fig:real_xy_traces_all_episodes}). The goal is to drive the vehicle at the handling limit while respecting a strict out-of-track (OOT) constraint. At deployment stage $k$, the agent is initialized from synthetic physics rollouts, Assetto Corsa data, and real-car trajectories collected before stage $k$. The resulting checkpoint is deployed using the same offline/online mixing protocol as in simulation.
Data collected during stage $k$ is used for online fine-tuning during that stage and is added to the offline buffer for subsequent stages. The initial 11 minutes of real-car seed data were collected using predefined maneuvers, including full-throttle acceleration and circular driving at different speeds. We use a deterministic MPC controller as a performance reference~\cite{lew2024riskaversemodelpredictivecontrol}.
We measure episodes to first success, success rate, best valid time under the OOT, average maximum OOT distance, and average speed. Failure-mode counts and adaptation progress are shown in Figure~\ref{fig:real_episode_outcome_counts}.

\begin{table}[t]
\centering
\caption{\textbf{Progressive real-car adaptation.} 
Each deployment stage is initialized from simulation priors plus accumulated real-car data collected before that stage. 
\emph{Real data} is the amount of real-car data (HH:MM) available before that stage; \textit{Real \%} is its percentage of the pretraining buffer; \emph{Succ. \%} is the episode index of the first episode classified as successful; \emph{Best time} is the fastest valid sector time with max $OOT<1m$ and no Did-not-finish (DNF).}
\label{tab:real_adaptation_results}
\footnotesize
\setlength{\tabcolsep}{0pt}
\renewcommand{\arraystretch}{1.05}
\begin{tabular*}{\columnwidth}{@{\extracolsep{\fill}}lccccccc@{}}
\toprule
Run & \makecell{Real\\data} & \makecell{Real\\\%} &
\makecell{Ep. to\\1st\\success} &
\makecell{Succ.\\\%} &
\makecell{Best time\\(s)\\OOT (m)} & \makecell{Mean\\max OOT\\(m)} &
\makecell{Mean\\speed\\(m/s)} \\
\midrule
\texttt{s1} & 00:11 & 0.28 & 5 & \mbox{37.5 (3/8)} & FAIL (--) & \mbox{1.53\,$\pm$\,0.09} & \mbox{10.99\,$\pm$\,3.68} \\
\texttt{s2} & 00:18 & 0.46 & 2 & \mbox{62.5 (5/8)} & 51.43 (0.13) & \mbox{0.30\,$\pm$\,0.29} & \mbox{14.63\,$\pm$\,0.98} \\
\texttt{s3} & 00:39 & 1.02 & \textbf{1} & \mbox{75.0 (6/8)} &
50.18 (0.17) & \mbox{\textbf{0.26\,$\boldsymbol{\pm}$\,0.11}} & \mbox{13.57\,$\pm$\,5.63} \\
\texttt{s4} & 00:44 & 1.13 & \textbf{1} & \mbox{\textbf{100 (8/8)}} &
\textbf{49.55} (0.30) & \mbox{0.32\,$\pm$\,0.29} & \mbox{\textbf{15.38\,$\boldsymbol{\pm}$\,0.94}} \\
\midrule
MPC & -- & -- & 1 & 100.0 & 52.47 (0.87) & 0.87 & 14.30 \\
\bottomrule
\end{tabular*}
\end{table}

Table~\ref{tab:real_adaptation_results} shows a clear real-car adaptation progression. 
With 11 minutes of accumulated real-car data, \texttt{s1} does not yet produce a valid OOT-constrained time and succeeds in only 37.5\% of evaluation episodes.
After adding additional real-car data, \texttt{s2} achieves its first success in episode 2, compared with episode 5 for \texttt{s1}, and reaches a valid time comparable to the baseline MPC. Stages \texttt{s3} and \texttt{s4} succeed on their first episode; by \texttt{s4}, the policy reaches 100\% success and improves both the best valid time and average speed. The overall progression shows that just a small 45-minute real-car anchor is sufficient to align the synthetic data replay buffer for reliable physical deployment. Figure~\ref{fig:real_episode_outcome_counts} shows how the outcome distribution changes across deployment stages. As more real-car data is added, failures decrease, and successful evaluations become more frequent. In addition, we tested PIA using only real-car data, without simulation priors. Deployment exhibited erratic control and was terminated for safety before completion. This highlights the practical value of PIA: under a limited real-data budget, simulation priors provide the structured initialization needed to make few-shot adaptation feasible.

\vspace{-0.08in}

\section{Discussion and Conclusion}
Simulation priors provide a practical offline initialization for model-based RL under strict target-interaction budgets. In ACGym, PIA-AC transfers to held-out track geometry, while target-geometry ODE trajectories further accelerate adaptation under high-fidelity dynamics and grip mismatch. ODE-only pretraining is useful, but the strongest results arise when low-fidelity coverage complements high-fidelity source data. On hardware, the same simulation-pretrained world-model pipeline supports progressive online adaptation on a full-size autonomous race car. This full-scale demonstration establishes that the approach can operate under real-time and physical-safety constraints. The central result is not only final lap performance, but that imperfect simulation priors enable faster and safer adaptation as limited target data are collected.

\section{Limitations}
PIA assumes that a model or simulator exists, and that synthetic and target data can be represented with a shared observation/action interface, making it more challenging to use for methods using complex sensor modes (image/LiDAR) as inputs, as this must be produced by the offline simulation environment. Our current experiments focus on settings where dynamics mismatch is the main challenge, not settings with strong exogenous stochasticity, such as opponents on track. The real-car study is limited to deployments on a single physical track and surface. The observed failure modes are: early real-car deployments can fail when the real-data anchor is too small, and safety-envelope violations remain possible during adaptation. 

\raggedbottom
\section*{APPENDIX}
\vspace{-0.1in}

\begin{table}[H]
\centering
\setlength{\tabcolsep}{2.0pt}
\renewcommand{\arraystretch}{1.15}
\caption{\textbf{Synthetic-only pretraining.} Pretraining on synthetic rollouts alone enables competitive adaptation. Adding a small amount of AC data improves over ODE-only, while the best lap time and safety are achieved by LOW ODE, where a small amount of ODE data complements a larger AC buffer.}
\footnotesize
\begin{tabular}{@{}lcccccc@{}}
\toprule
Model &
\makecell{ODE\\(M)} &
\makecell{AC\\(M)} &
\makecell{Best lap\\(s)} &
\makecell{\# Ep. to\\1st success} &
\# Failures &
\makecell{Miles per\\Failure} \\
\midrule
SCRATCH & 0.0 & 0.0 & Fail & -- & 50 & 0.03 \\
ODE ONLY & 39.0 & 0.0 & 115.90 & 36 & 46 & 3.98 \\
AC ONLY & 0.0 & 36.0 & 114.16 & 8 & 18 & 34.62 \\
LOW AC & 39.0 & 1.0 & 112.90 & 6 & 19 & 33.46 \\
LOW ODE & 3.0 & 36.0 & \textbf{112.17} & \textbf{5} & \textbf{11} & \textbf{68.80} \\
\bottomrule
\end{tabular}
\label{tab:synthetic_only_pretraining}
\end{table}
\vspace{-0.2in}

\begin{table}[H]
\centering
\setlength{\tabcolsep}{5.0pt}
\renewcommand{\arraystretch}{1.15}
\caption{\textbf{Ablations.} ODE data (partitioning \textbf{G}eometry and \textbf{D}ynamics), sub-task ID, and uncertainty penalization.}
\begin{tabular}{lccc}
\toprule
Experiment & Best Lap (s) &
\# Failures & \makecell{Miles per\\Incident} \\ \midrule
No ODE & 114.16 & 18 & 34.62 \\
ODE G only  & 113.39 & 14 & 49.72 \\
ODE D only  & 114.43 & 27 & 17.46 \\
ODE G+D     & \textbf{112.17} & \textbf{11} & \textbf{68.80} \\
\midrule
W/o context & 112.98 & 29 & 19.12 \\
W/ context  & \textbf{112.17} & \textbf{11} & \textbf{68.80} \\
\midrule
W/o uncertainty in~\eqref{eq:regularized_planning_score} & 114.54 & 20 & 8.32 \\
W/ uncertainty in~\eqref{eq:regularized_planning_score}  & \textbf{113.16} & \textbf{9} & \textbf{37.62} \\
\bottomrule
\end{tabular}
\label{tab:ablations_merged}
\end{table}
\vspace{-0.22in}

\begin{table}[H]
\centering
\setlength{\tabcolsep}{4pt}
\caption{\textbf{Domain-randomized ODE priors.} All methods are fine-tuned and evaluated on the Silverstone track.}
\begin{tabular}{lcc}
\toprule
Offline data & Best lap (s) $\downarrow$ & Miles per Failure $\uparrow$ \\
\midrule
$\mathrm{ODE}_{\mathrm{DR,SIL}}+\mathrm{AC}_{\mathrm{BRN}}$ & 113.37 & \textbf{13.31} \\
$\mathrm{ODE}_{\mathrm{SIL}}+\mathrm{AC}_{\mathrm{BRN}}$ & 113.37 & 11.26 \\
$\mathrm{AC}_{\mathrm{BRN}}$ & 115.46 & 5.20 \\
$\mathrm{ODE}_{\mathrm{DR,SIL}}$ & 115.54 & 3.43 \\
$\mathrm{ODE}_{\mathrm{DR,BRN}}$ & 117.01 & 2.62 \\
\bottomrule
\end{tabular}
\label{tab:ode_dr}
\end{table}
\vspace{-0.15in}

\subsubsection{Synthetic-only pretraining}
\label{sec:appendix:synthetic_only}

We evaluate whether synthetic ODE rollouts can support adaptation when AC data is limited. This experiment asks whether policies pretrained exclusively on synthetic rollouts can achieve competitive adaptation, and how ODE and AC data trade off under different pretraining mixtures. We compare \textbf{SCRATCH} with no pretraining, \textbf{ODE ONLY} pretrained solely on ODE rollouts, \textbf{AC ONLY} pretrained solely on simulator rollouts, \textbf{LOW AC} with 39M ODE steps and 1M AC steps, and \textbf{LOW ODE} with 3M ODE steps and 36M AC steps. Table~\ref{tab:synthetic_only_pretraining} shows that synthetic-only pretraining is viable when high-fidelity data is unavailable, but low-fidelity data is most effective as complementary coverage rather than as a full replacement.

\subsubsection{Ablations}
\paragraph{ODE Signal ablations}
\label{sec:appendix:ode_signals}
We seek to determine whether ODE pretraining helps via geometry (spatial layout: track borders, curvature look-ahead), dynamics (proprioceptive signals), or both. We compare pretraining settings to isolate their contributions. We pretrain with: No-ODE (PIA-MBRL-AC), ODE (G+D) (PIA-ODE), ODE G-only (mask dynamics on ODE samples), ODE D-only (mask geometry on ODE samples) and fine-tune on SIL. Geometry-only reduces failures relative to No-ODE, confirming that spatial priors improve exploration safety, but has a slower lap time. Dynamics-only underperforms on both metrics, while G+D is best on both.
\paragraph{Planning Uncertainty Penalty}
We ablate the planner regularization mechanism described in Section~\ref{sec:method}. Removing it leads to unstable adaptation and more incidents. Regularization stabilizes early adaptation.
\paragraph{Context Ablation}
We ablate conditioning on the context $c$. Removing the context leads to worse
generalization and noisier adaptation, confirming that explicit context helps
disambiguate heterogeneous datasets.
\paragraph{Domain-Randomized ODE Prior}
\label{sec:appendix:ode_dr}
We additionally test whether domain randomization can replace or complement heterogeneous simulation priors. We randomize tire grip, vehicle mass, engine torque, and yaw inertia in the ODE generator, then fine-tune and evaluate on SIL. We observe that Domain randomization alone does not improve ODE-only pretraining. When combined with the same AC data, however, it preserves the best lap time and improves miles per failure. These results indicate that domain randomization is complementary to, rather than a replacement for, combining heterogeneous priors with target-domain adaptation.

\bibliographystyle{IEEEtran}
{\footnotesize
\bibliography{main}
}

\end{document}

%% file: tikz/mosaic.tex
\newlength{\imagewidth}
\newlength{\hpadding}
\newlength{\vpadding}
\setlength{\imagewidth}{3.3cm}
\setlength{\hpadding}{0.5mm}
\setlength{\vpadding}{0.5mm}
\newcommand\flip[1]{\scalebox{-1}[1]{#1}}

\begin{tikzpicture}[inner sep=0pt,outer sep=0pt]
  \node[anchor=north west] (good1) at (0,0)
    {\flip{\includegraphics[width=\imagewidth]
     {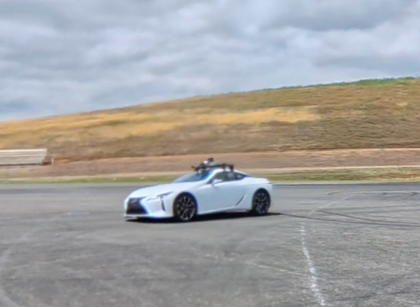}}};

  \foreach \i [evaluate=\i as \prev using int(\i-1)] in {2,...,5} {
    \node[anchor=north west] (good\i)
      at ([xshift=\hpadding]good\prev.north east)
      {\flip{\includegraphics[width=\imagewidth]
       {tikz/bad/frame_0000000\i.png}}};
  }

  \node[anchor=north west] (bad1)
    at ([yshift=-\vpadding]good1.south west)
    {\flip{\includegraphics[width=\imagewidth]
     {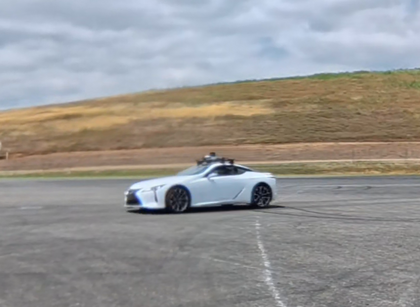}}};

  \foreach \i [evaluate=\i as \prev using int(\i-1)] in {2,...,5} {
    \node[anchor=north west] (bad\i)
      at ([xshift=\hpadding]bad\prev.north east)
      {\flip{\includegraphics[width=\imagewidth]
       {tikz/good/frame_0000000\i.png}}};
  }
  
    \foreach \i [evaluate=\i as \prev using int(\i-1)] in {1,...,5} {
    \node[anchor=north] (llbl\i)
      at ([yshift=0.3cm, xshift=\hpadding]good\i.north) {$t_\prev$};
  }
  
  \node[anchor=south, rotate=90] (A1) at ([xshift=-0.1cm]good1.west)
    {\begin{tabular}{c}\small Before\\ \small Adaptation\end{tabular}};
  \node[anchor=south, rotate=90] (A2) at ([xshift=-0.1cm]bad1.west)
    {\begin{tabular}{c}\small After\\ \small Adaptation\end{tabular}};
      
\end{tikzpicture}